\documentclass[runningheads]{llncs}
\usepackage[T1]{fontenc}
\usepackage{graphicx}
\usepackage{booktabs}
\usepackage{tabularx}
\usepackage{amsmath}
\usepackage{amsfonts}
\usepackage{amssymb}
\usepackage{hyperref}
\usepackage{xcolor}

\begin{document}
\title{SPiKE: 3D Human Pose\\from Point Cloud Sequences}

\author{Irene Ballester \inst{1}\orcidID{0000-0002-0219-9063}\and Ondřej Peterka \inst{1}\orcidID{0009-0002-7231-0430} \and Martin Kampel \inst{1}\orcidID{0000-0002-5217-2854}}
\authorrunning{Ballester, I., Peterka, O. and Kampel, M.}

\institute{Computer Vision Lab, TU Wien \\ \email{\{irene.ballester,martin.kampel\}@tuwien.ac.at}, \email{ondra.peterka@gmail.com}}

\maketitle              
\begin{abstract}

3D Human Pose Estimation (HPE) is the task of locating keypoints of the human body in 3D space from 2D or 3D representations such as RGB images, depth maps or point clouds. Current HPE methods from depth and point clouds predominantly rely on single-frame estimation and do not exploit temporal information from sequences. This paper presents SPiKE, a novel approach to 3D HPE using point cloud sequences. Unlike existing methods that process frames of a sequence independently, SPiKE leverages temporal context by adopting a Transformer architecture to encode spatio-temporal relationships between points across the sequence. By partitioning the point cloud into local volumes and using spatial feature extraction via point spatial convolution, SPiKE ensures efficient processing by the Transformer while preserving spatial integrity per timestamp. Experiments on the ITOP benchmark for 3D HPE show that SPiKE reaches 89.19\% mAP, achieving state-of-the-art performance with significantly lower inference times. Extensive ablations further validate the effectiveness of sequence exploitation and our algorithmic choices. 
Code and models are available at: \url{https://github.com/iballester/SPiKE}
\end{abstract}
\section{Introduction} 

3D Human Pose Estimation (HPE) aims to localize body keypoints or joints in the 3-dimensional space from images, videos and 3D representations, such as point clouds. This task faces significant challenges due to the occlusion of body parts, diversity in human postures, and the wide range of human appearances and shapes. Achieving precise joint localisation is critical for numerous applications in the real world~\cite{zheng2023deep}, including human activity recognition~\cite{sun2022human}, gait analysis~\cite{gu2021cross,teepe2022towards} and motion forecasting~\cite{diller2022forecasting}.

Methods for HPE from RGB have received more attention than depth-based approaches~\cite{jeong2023solopose,zhao2023poseformerv2,zhu2023motionbert}. However, the depth modality offers distinct advantages for 3D HPE, encoding inherent 3D information and exhibiting robustness to diverse lighting conditions while preserving privacy better than RGB~\cite{ballester2024action,mucha2022addressing}. Depth maps serve as a representation of 3D space in a 2D image, allowing the direct extraction of 3D HPE from a single 2D depth map~\cite{garau2021deca,xiong2019a2j}. However, using 2D depth maps encounters challenges such as perspective distortion and a non-linear mapping between depth maps and 3D coordinates, which hinder the learning process~\cite{moon2018v2v,zhou2020learning}. To overcome this limitation, 3D representations are derived from depth maps, such as voxels~\cite{moon2018v2v} or point clouds~\cite{zhou2020learning}, facilitating the \textcolor{black}{direct} estimation of 3D keypoints. In line with these works, we employ point clouds from depth maps as input for our method.

\begin{figure}[t]
    \centering
    \includegraphics[width=\linewidth]{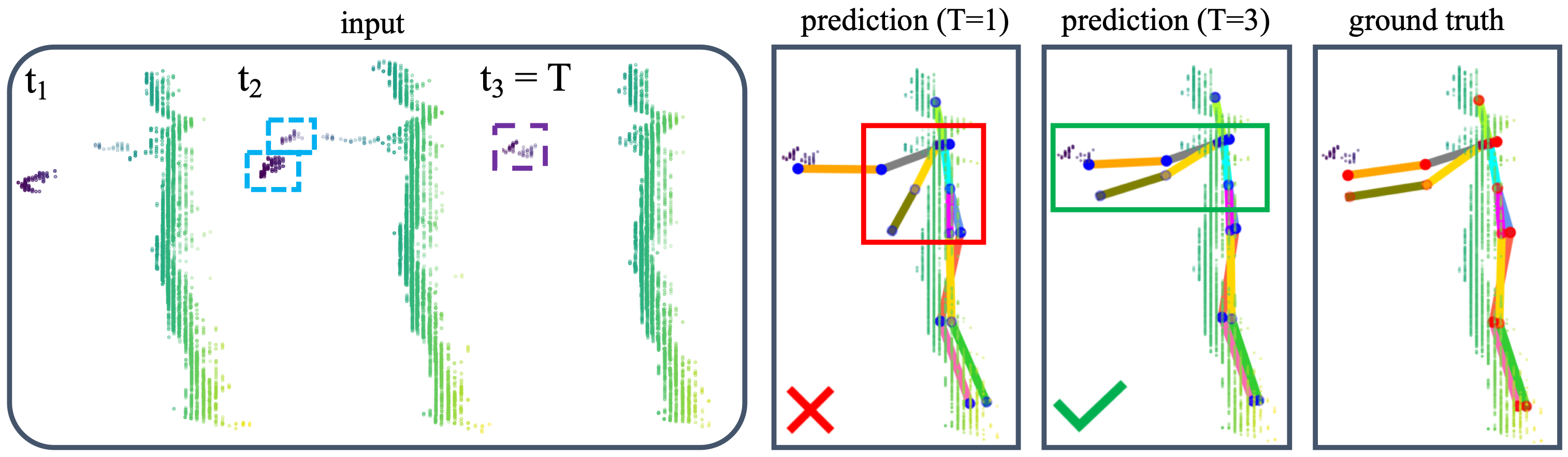} %
    \caption{\textbf{Importance of exploiting sequence information}. When considering only the current frame (sequence length $T$=1), only the right hand is visible in the input point cloud, leading to an incorrect prediction. On the contrary, if we consider past frames ($T$=3), in particular $t_2$ where both hands are visible, SPiKE estimates the position of both arms more accurately. Timestamp ID: 3\_02244.}
    \label{fig:teaser}
\end{figure}

Different from direct methods, 2D-3D lifting methods~\cite{DEusanio2023,zhang2020weakly,zhang2021sequential}, inspired by lifting approaches in RGB~\cite{mehraban2024motionagformer,zhao2023poseformerv2,zhu2023motionbert}, first estimate the 2D keypoints from the depth map. Then, they use the z-coordinate of the estimated 2D keypoints in the depth map to project its coordinates into 3D space and refine the 3D prediction, typically using a point cloud representation. This comes at the cost of increased complexity, as methods must deal with more than one input type. Furthermore, since depth-based methods rely on dense maps to extract the coordinates of the joints, they are not compatible with sparse point clouds~\cite{zhang2021sequential} and would require additional algorithms to densify the point cloud~\cite{cui2021deep,uhrig2017sparsity}. In this work, we propose a pure point cloud method, that achieves state-of-the-art performance with a reduced inference time.

Most of the presented works process the timestamps of sequences of depth maps or 3D representations independently without using sequence information and thus without taking advantage of temporal relationships between frames. In the RGB domain, sequence information proves beneficial for pose estimation to cope with occlusion~\cite{jeong2023solopose,wen2023hierarchical}. Inspired by recent advances in dynamic point cloud processing~\cite{fan2021point,fan2022point,wen2022point}, we propose to use a Transformer~\cite{vaswani2017attention} to process sequential point clouds for 3D HPE. As illustrated in Fig. \ref{fig:teaser}, sequence information makes the model more robust against occluded body parts and noise.

More specifically, we divide each point cloud of the sequence into local volumes and extract spatial features within them. The input tokens for the Transformer are generated by combining the 4D coordinates of the centroid of the local volumes and the features of that volume extracted by a point convolution. This is fed into a Transformer~\cite{vaswani2017attention} that performs global self-attention to encode the spatio-temporal relationships of the points along the sequence to predict the 3D coordinates of the joints. Our method, SPiKE (Sequential Point clouds for Keypoint Estimation), is validated on the ITOP dataset~\cite{Haque2016} and outperforms the state of the art, confirming its suitability for 3D HPE from sequential point clouds. Our contributions are as follows:

\begin{itemize}
    \item We introduce SPiKE, a novel approach for 3D HPE from point cloud sequences. Unlike previous works that process timestamps independently, our method leverages temporal information by employing a Transformer to encode the spatio-temporal structure along the sequence. To ensure efficient processing by the Transformer while preserving spatial integrity per frame, SPiKE partitions each point cloud of the sequence into local volumes for feature extraction through point spatial convolution.. 
    \item Experiments, qualitative results and comparisons with the state of the art confirm the effectiveness of our approach. SPiKE achieves an mAP of 89.19\% for 3D HPE on the ITOP dataset~\cite{Haque2016}, outperforming existing \textcolor{black}{direct} models from depth maps, voxels or point clouds. Furthermore, our model performs similarly to \textcolor{black}{2D-3D lifting approaches with a significantly lower inference time since no depth-branch is required.} 
    \item Extensive ablation studies confirm the value of leveraging sequence information, retaining spatial structure per timestamp, and our algorithmic choices.

\end{itemize}

\section{Related work}

\subsection{Human Pose Estimation from 3D Information}

\subsubsection{Direct methods for 3D HPE: depth maps, voxels and point clouds} The depth modality is different from RGB in that, by its nature, it already contains 3D information in its 2D form. Early works in HPE from depth extract the 3D coordinates directly from the 2D depth map~\cite{carreira2016human,pavlakos2017coarse,wang2016human,wang2018convolutional,xiong2019a2j}. Arguing a lack of generalisation capabilities between different perspectives, DECA~\cite{garau2021deca} utilizes Capsule Networks~\cite{hinton2018matrix} to model inherent geometric relations in human skeletons to achieve viewpoint-equivariance. Depth maps offer the advantage of lightweight data storage and processing, and their 2D nature facilitates the adaptation of RGB models and pre-trained feature extractors, broadening the scope of available methods and datasets. However, they suffer from perspective distortion~\cite{moon2018v2v,wang20203dv}. To overcome this limitation, Moon et al.~\cite{moon2018v2v} propose to voxelise the depth map to obtain a volumetric representation and generate per-voxel likelihoods for each keypoint. Despite effectively solving the problem of perspective distortion, voxels also present challenges in terms of computational demands and unavoidable quantisation errors during voxelisation, which is particularly relevant for HPE where precise scene geometry measurement is crucial.

Point clouds require memory relative to the number of points and provide arbitrary precision. In this line, Zhou et al.~\cite{zhou2020learning}, adapt stacked EdgeConv layers from DGCNN~\cite{wang2019dynamic} and T-Net from PointNet~\cite{qi2017pointnet} to regress the 3D positions of the joints. More recently, Weng et al.~\cite{weng20233d} propose an unsupervised pre-training strategy for 3D HPE from point clouds. LiDAR-HMR \cite{fan2023lidar} estimates 3D human body mesh from sparse point clouds by first estimating the 3D human pose to then employ a sparse-to-dense 3D mesh reconstruction approach. LPFormer~\cite{ye2024lpformer} proposes a top-down multitask approach for 3D HPE from sparse point clouds.

Despite the advances in 3D HPE from depth maps and point clouds resulting from the approaches presented, they all process each depth map or point cloud independently and cannot directly process sequences due to the lack of an inter-frame feature fusion approach. Encouraged by the success of integrating sequence information in the RGB domain~\cite{arnab2019exploiting,jeong2023solopose,wen2023hierarchical}, we propose the use of a Transformer architecture~\cite{vaswani2017attention} to encode sequence information.

\subsubsection{2D-3D lifting models for 3D HPE: depth maps + point clouds}
Extracting an intermediate 2D pose from depth maps and then refining its 3D projection with point clouds proves to be an effective strategy. In this line, inspired by RGB 2D-3D lifting methods, D'Eusanio et al.~\cite{DEusanio2023} evaluate the modular refinement network RefiNet~\cite{d2021refinet} using as starting point 2D keypoints from HRNet~\cite{wang2019dynamic}. Following a similar strategy for hand pose estimation, Ren et al.~\cite{ren2023} iteratively correct the 3D projection of the estimated 2D hand keypoints by taking a feature set from a local region around each estimated joint. Zhang et al.~\cite{zhang2020weakly} use depth maps to obtain an intermediate 2D pose estimate and sampled point cloud, and then refine the estimates by processing point clouds through PointNet~\cite{qi2017pointnet}. An ablation study presented in this work shows that using 2D predictions as a starting point instead of direct 3D estimation from point clouds improves the overall accuracy by almost 14 points on the ITOP dataset, demonstrating that combining these modalities is an effective strategy. Building on this work, Adapose~\cite{zhang2021sequential} adds to this pipeline 1) an adaptive sampling strategy for point clouds and 2) an LSTM module to capture inter-frame features and enforce temporal smoothness, demonstrating the benefits of using sequential point cloud processing. This last finding, coupled with the evidence from the RGB modality, further strengthens our argument for the use of sequence information. 

One of the main contributions is that SPiKE takes only point clouds as input, without requiring depth maps for intermediate 2D estimation. This not only provides versatility by allowing seamless integration with different point cloud acquisition methods, but also significantly reduces inference time as SPiKE performs direct estimation in 3D.

\subsection{Deep learning for dynamic point clouds}

Point cloud sequences, unlike grid-based RGB video, lack regularity in spatial arrangement as points appear inconsistently over time. One approach to address this lack of order is to voxelise the 3D space and apply 4D grid-based convolutions. In this line, Choy et al.~\cite{choy20194d} extend the temporal dimension of 3D sparse convolutions~\cite{graham2017submanifold} to extract spatio-temporal features on 4D occupancy grids. 3DV~\cite{wang20203dv} combines voxel-based and point-based modelling by first integrating 3D motion information into a regular compact voxel set and then applying PointNet++~\cite{qi2017pointnet++} to extract representations via temporal rank pooling~\cite{fernando2016rank}.

An alternative to voxelisation is to operate directly on point sets, avoiding the quantisation errors inherent in the voxelisation process. In this line, MeteorNet~\cite{liu2019meteornet} extends PointNet++~\cite{qi2017pointnet++} for 4D point cloud processing to collect information from neighbours and relies on point tracking to merge points across timestamps. PSTNet~\cite{fan2021pstnet} decomposes spatial and temporal data and proposes a hierarchical point-based convolution. To avoid point tracking, P4Transformer~\cite{fan2021point} proposes to use a Transformer architecture~\cite{devlin2018bert,vaswani2017attention} to perform self-attention over the whole sequence after encoding spatio-temporal local regions by a 4D point convolution. PST-Transformer~\cite{fan2022point} modifies the Transformer architecture of \cite{fan2021point} to preserve the spatio-temporal encoding structure. 

These methods are effective in downstream tasks such as semantic segmentation and activity recognition. In this work, we show that point convolutions and attention-based architectures are also suitable for HPE. Similarly to~\cite{fan2021point,fan2022point}, we use a Transformer to relate local volumes from different timestamps. However, different from~\cite{fan2021point,fan2022point}, we propose to use spatial local regions instead of spatio-temporal ones. The rationale behind this is that while temporal merging is suitable for action recognition, allowing for longer sequences that yield better performance, this strategy is not directly applicable to HPE. For HPE, we show that while sequence information is beneficial, longer sequences beyond a certain length do not improve performance (consistent with~\cite{wen2023hierarchical} in RGB). This choice preserves spatial structure by merging only spatial information within the same timestamp before passing local features to the Transformer.

\section{Method}

SPiKE processes a sequence of point clouds, \([P_1, P_2, \dots, P_T]\), each containing \(N\)~randomly sampled points that are represented as \(P_t = \{p_i(t)\}_{i=1}^N\) for timestamp~\(t\) within total sequence length \(T\). Each point \(p_i(t)\) is defined by its Euclidean coordinates in \(\mathbb{R}^3\). Our goal is to predict the 3D locations of \(M\) key body joints, represented as \(J = [j_1, j_2, \dots, j_M]\).

\begin{figure}[t]
    \centering
    \includegraphics[width=\linewidth]{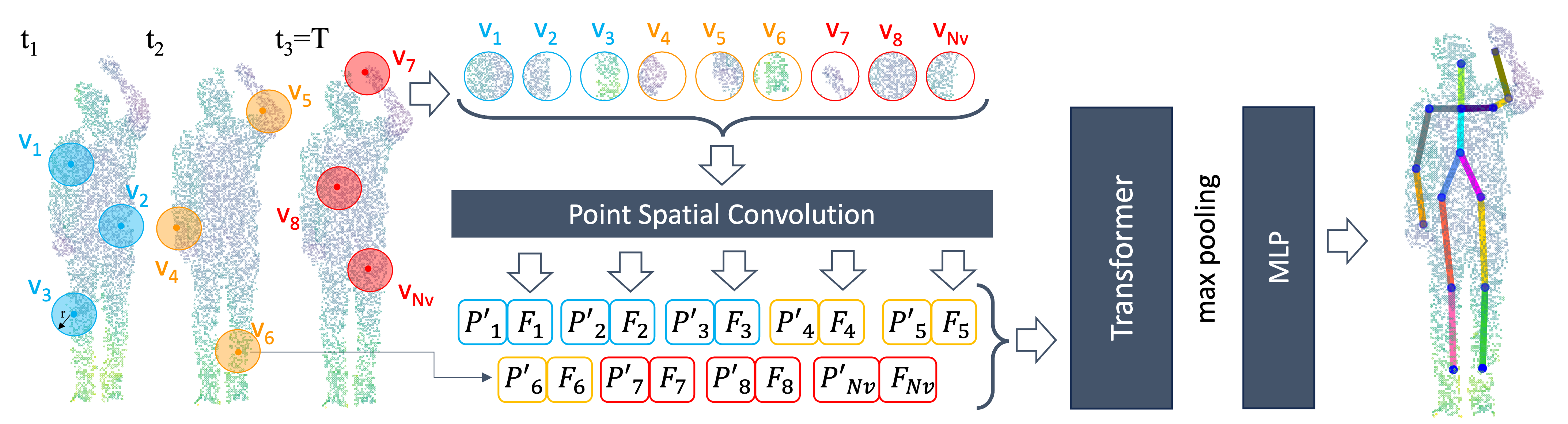} %
    \caption{\textbf{SPiKE pipeline.} First, each point cloud of the sequence (sequence length~=~$T$) is partitioned by selecting $N_v$ reference points $P'_i$ and creating local volumes $V_{N_v}$ around them by sampling points within a radius $r$. Point Spatial Convolution extracts spatial features $F_i$ from each local volume. These features are then embedded with the coordinates of their respective reference point $P'_i$  and fed into the Transformer. After a max-pooling layer, an MLP regresses the 3D coordinates of the $M$ joints.}
    \label{fig:pipeline}
\end{figure}

The entire pipeline is illustrated in Fig. \ref{fig:pipeline}. First, we extract spatial features by applying a point spatial convolution to local regions in the point cloud of each timestamp (described in Section \ref{sec:conv}). These local features, together with a positional encoding, are then processed by a Transformer architecture that merges information across different timestamps (described in Section \ref{sec:transf}). Subsequently, a max-pooling operation merges the transformed local features into a single global feature representation. Finally, a multilayer perceptron (MLP) regresses the 3D coordinates of the $M$ joints.

\subsection{Point Spatial Convolution in local regions}
\label{sec:conv}

Point cloud sequences represent dynamic 3D environments with a large number of individual points. The direct application of self-attention across all points in such sequences proves to be computationally expensive and demanding in terms of running memory. Following~\cite{fan2021point,fan2021pstnet,qi2017pointnet++}, we construct $N_v$ local regions (hereinafter referred to as local volumes) and perform a point spatial convolution to encode the local structure of the points within these volumes.

Each point cloud is divided into $N_v$ local volumes  ${V_1, V_2, ..., V_{N_v}}$ centred around reference points. The selection of reference points ${P'_1, P'_2, ..., P'_{N_v}}$ is carried out using Farthest Point Sampling (FPS)~\cite{qi2017pointnet++}, ensuring that these points are strategically distributed across the point cloud. For each reference point $P'_i$, we then sample $N_s$ neighboring points within a radius $r$, again using FPS.

After creating the local volumes, we apply a point spatial convolution to encode the spatial relationships among the $N_s$ neighboring points contained within. This process transforms the original point cloud sequence, designated as $[P_1, P_2, ..., P_T]$, with $P_t \in \mathbb{R}^{3 \times N}$ representing the set of point coordinates at the $t$-th frame into a sequence of encoded features. Each timestamp in the transformed sequence is represented as $\left[ P'_1; F_1 \right], \left[ P'_2; F_2 \right], ..., \left[ P'_T; F_T \right]$,  where $P'_t~\in~\mathbb{R}^{3 \times N_v}$ and $F_t\in \mathbb{R}^{C\times N_v}$, with C denoting the number of feature channels. For any given reference point $P'_i$, located at $(x, y, z, t)$, its feature vector $F(x,y,z,t) \in \mathbb{R}^{C \times 1}$ is derived from the spatial convolution as follows:

\begin{equation}
F(x,y,z,t) = \max_{\substack{\|(\delta_x, \delta_y, \delta_z)\| \leq r
}}(\text{MLP}(W_s\cdot (\delta_x, \delta_y, \delta_z)^T))
\end{equation}

Here, $W_s \in \mathbb{R}^{C' \times 3}$ represents the transformation matrix applied to the 3D displacements $(\delta_x, \delta_y, \delta_z)$, encapsulating the spatial differences relative to the reference point. This matrix multiplication facilitates the projection of spatial displacements into a higher-dimensional feature space, which is subsequently processed by an MLP to enhance the representation. Finally, we aggregate the features by performing max pooling within the local region.

\subsection{Transformer}
\label{sec:transf}

\subsubsection{Positional embedding}
After the point spatial convolution, the local volumes of the $t$-th frame are encoded to features $F(x,y,z,t)$. These features, however, solely capture the local spatial features without explicitly accounting for the absolute positions of the reference points within the global structure of the point cloud. To address this limitation and ensure the preservation of the spatio-temporal structure inherent to the point cloud sequence, we combine the coordinates of the reference point, i.e., $P'(x, y, z, t)$, and local area features as input to the Transformer.

\begin{equation}
I{(x,y,z,t)} = W_i \cdot P'(x, y, z, t)^T + F(x,y,z,t)
\end{equation}

In this equation, $W_i \in \mathbb{R}^{C \times 4}$ represents a weight matrix that transforms the four-dimensional coordinates $P'(x, y, z, t)$ into a feature space that is compatible with the encoded local features $F(x,y,z,t)$. The result of this transformation, $I{(x,y,z,t)}$, serves as the input to the Transformer, where $I \in \mathbb{R}^{C \times TN_v}$ are the transformed input features ready for further processing. By embedding the spatial and temporal coordinates directly into the feature representation, the subsequent Transformer layer can take advantage of both the local feature information and the positional context of each reference point. 

\subsubsection{Multi-head self-attention}

The multi-head self-attention mechanism~\cite{vaswani2017attention} enables the model to capture spatial and temporal dependencies within the sequences of point clouds. Input features $I{(x,y,z,t)} \in \mathbb{R}^{C \times TN_v}$ representing local spatial features and positional embeddings are transformed into query ($Q$), key ($K$), and value ($V$) matrices through linear transformations. Specifically, for each local volume centered at $P'(x, y, z)$ at timestamp $t$, we compute:
\[
Q = I{(x, y, z, t)} \cdot W_Q, \quad K = I{(x, y, z, t)} \cdot W_K, \quad V = I{(x, y, z, t)} \cdot W_V,
\]

where \( W_Q \in \mathbb{R}^{C_k \times C}\), \( W_K \in \mathbb{R}^{C_k \times C}, \) \( W_v \in \mathbb{R}^{C_v \times C} \) are learnable weight matrices, and \( C_k \) and \( C_v \) are the dimensions of key and value, respectively.

The attention mechanism computes attention scores based on the similarity between queries and keys, determining the relevance of different spatial positions and timestamps within the input sequence. For each local volume centered at $P'(x, y, z)$ at timestamp $t$, the attention scores are calculated as follows:

\begin{equation}
\text{Attention}(Q,K,V) = \text{softmax}\left(\frac{QK^T}{\sqrt{d_k}}\right)V
\end{equation}

where $d_k$ represents the dimensionality of the key vectors. The softmax function normalizes the attention scores across the key vectors, indicating the importance of each value vector relative to the given query.

To capture diverse patterns and dependencies within the data, the multi-head mechanism splits the query, key, and value matrices into $h$ separate heads, each operating independently. This parallel processing enables the model to attend to different parts of the input simultaneously, enhancing its ability to capture both local and global dependencies. The outputs of the individual heads are then concatenated and linearly transformed by $W_o$, resulting in the final output of the multi-head self-attention mechanism. Finally, the model incorporates $m$ Transformer blocks, each equipped with a multi-head self-attention mechanism.

\subsection{Implementation details}

SPiKE is trained end-to-end with L1 loss and SGD optimizer (batch size = 24, learning rate = 0.01) for 150 epochs. We use 4096 randomly sampled points, with $r$ = 0.2, $N_v$ = 128, and $N_s$ = 32. Furthermore, $C$ = 1024, and the Transformer has $m$ = 5 self-attention blocks with $h$ = 8 heads each. Point clouds are centred by subtracting the mean of 3D coordinates of the points per sequence, with rotation in the y-axis of [-90, 90] degrees and x-axis mirroring for augmentation. Training and testing are performed on a single NVIDIA GeForce RTX 3090.

To isolate the points belonging to the human, following~\cite{zhou2020learning}, we use depth thresholding to remove the background and discard the first 10 bins of the y-coordinate histogram to exclude floor points. Then, clusters are formed using DBSCAN~\cite{dbscan} with a 15 cm inter-cluster distance. Since humans may not always form a single cluster, we select the largest cluster and include clusters below, above and between the largest cluster and the sensor, offset by 20 cm.

\section{Evaluation}

We describe the dataset and evaluation metrics, and systematically evaluate SPiKE by comparing it to the state of the art, discussing qualitative results, and providing ablations to illustrate our contributions.

\subsection{Datasets and metrics}

\subsubsection{ITOP Human Pose Dataset~\cite{Haque2016}}is a collection of 100k depth maps from two camera viewpoints captured with Asus Xtion Pro sensors. It consists of 15 action sequences performed by 20 subjects. All depth maps are labelled with 3D coordinates of 15 body joints from the camera viewpoint. We train and test SPiKE on ITOP front-view and adopt the original division proposed in~\cite{Haque2016}, i.e., using subjects 00-04 for testing and subjects 05-19 for training, so that our evaluation reflects a scenario where testing is performed on unseen subjects. 

In ITOP, only about 45\% of the annotated joints are human-validated (referred to as ``valid joints''), and methods typically evaluate performance only on these valid joints~\cite{zhang2020weakly,zhang2021sequential,zhou2020learning}, hence, we train and test our method only on validated ground truth annotations. The point clouds of instances with invalid joints are incorporated into the sequence to predict subsequent valid joint positions, but the invalid joints are never used for training or testing.

\subsubsection{Mean Average Precision (mAP)}

Following previous work~\cite{DEusanio2023,garau2021deca,moon2018v2v,xiong2019a2j}, we use mean Average Precision (mAP) as the evaluation metric with a threshold of 10 cm. mAP is the percentage of all predicted joints that fall within an interval of less than 0.10 metres of the 3D coordinates of the ground truth joints.

\subsection{Comparison with state-of-the-art methods}

We evaluate SPiKE on ITOP front-view against state-of-the-art methods for 3D HPE on depth maps, point clouds and voxels in Table~\ref{tab:sota}. For better comparison, the different approaches are classified as direct and 2D-3D lifting methods (SPiKE belongs to the former). Specifically, we compare our approach with the following direct methods: V2V~\cite{moon2018v2v}, A2J~\cite{xiong2019a2j}, Zhou et al.~\cite{zhou2020learning} and DECA~\cite{garau2021deca}. For 2D-3D lifting methods, we consider WSM~\cite{zhang2020weakly}, AdaPose~\cite{zhang2021sequential}, and HRNet+ RefiNet~\cite{DEusanio2023}. For reference, we also include the ablation study by~\cite{zhang2020weakly} as ``WSMa'' as part of the direct methods, since in this ablation the 3D pose is estimated from the point clouds without relying on an intermediate extraction of 2D keypoints.

\begin{table}[h]
\centering
\small

\caption{Comparison with the state-of-the-art methods on ITOP front-view (0.1m mAP). \textcolor{black}{(*) identifies the methods using additional training data.}}
\begin{tabularx}{\textwidth}{l*{6}{>{\centering\arraybackslash}X}|*{3}{>{\centering\arraybackslash}p{1.1cm}}}
\toprule
 & \multicolumn{6}{c|}{\textcolor{black}{direct methods}} & \multicolumn{3}{c}{\textcolor{black}{2D-3D lifting methods}} \\
\cmidrule(lr){2-7} \cmidrule(lr){8-10}
Method & V2V & A2J & WSMa & Zhou et al. & DECA & SPiKE (Ours) & WSM\textcolor{black}{*}  & AdaPose & \multicolumn{1}{>{\centering\arraybackslash}p{1.5cm}}{HRNet+ RefiNet} \\
& 2018 & 2019 & 2020 & 2020 & 2021 & - & 2020 & 2021 & 2023 \\
\midrule
Modality & voxels & depth & points & points & depth & points & & depth+points &  \\
\toprule
Head & 98.29 & 98.54 & - & 96.73 & 93.87 & 98.42 & 98.15 & 98.42 & - \\
Neck & 99.07 & 99.20 & - & 98.05 & 97.90 & 99.47 & 99.47 & 98.67 & - \\
Shoulders & 97.18 & 96.23 & - & 94.38 & 95.22 & 97.48 & 94.69 & 95.39 & - \\
Elbows & 80.42 & 78.92 & - & 73.67 & 84.53 & 81.64 & 82.80 & 90.74 & - \\
Hands & 67.26 & 68.35 & - & 54.95 & 56.49 & 71.71 & 69.10 & 82.15 & - \\
Torso & 98.73 & 98.52 & - & 98.35 & 99.04 & 99.24 & 99.67 & 99.71 & - \\
Hips & 93.23 & 90.85 & - & 91.77 & 97.42 & 93.68 & 95.71 & 96.43 & - \\
Knee & 91.80 & 90.75 & - & 90.74 & 94.56 & 91.56 & 91.00 & 94.41 & - \\
Feet & 87.60 & 86.91 & - & 86.30 & 92.04 & 84.30 & 89.96 & 92.84 & - \\
\midrule

Upper B. & - & - & - & 80.10 &  83.03 & 88.75 & - & - & 80.8 \\
Lower B. & - & - & - & 89.60 & 95.30 & 89.85 & - & - & 88.1 \\
\midrule

Mean & 88.74 & 88.00 & 75.64 & 85.11 & 88.75 & \textbf{89.19} & 89.59 & \textbf{93.38} & 84.2 \\
\bottomrule
\label{tab:sota}
\end{tabularx}
\end{table}

SPiKE ($T=3$ and only past timestamps) achieves an overall mAP of 89.19\%, outperforming existing direct methods using any of the modalities: depth maps, points and voxels. Our method shows significant improvements, most notably for the upper limbs, which are prone to occlusion when the person is sideways or moving their arms. In these cases of occlusion, the sequence information becomes valuable as certain timestamps can reveal visible joints that are occluded at that particular timestamp, providing crucial context for accurate estimation.

Compared to direct methods working with point clouds, lifting methods can leverage 2D pre-trained backbones with additional data (marked with * in Table~\ref{tab:sota}). This is an effective strategy to improve performance, but prevents a direct and fair comparison with methods trained only in ITOP. Nevertheless, despite using additional data in the WSM 2D HPE network training, SPiKE (trained only on ITOP) achieves comparable performance with a difference of only 0.4 points. The reliance of 2D-3D lifting approaches on the estimation of an intermediate pose from the depth map is evident from the performance drop of 14 points between WSM (mAP = 89.59\%) and WSMa (mAP = 75.64\%) when no intermediate pose is considered. In standard WSM, intermediate 2D keypoints are first extracted, reprojected in 3D to obtain an intermediate 3D pose estimate, and then refined by processing the point clouds. In contrast, in the WSMa ablation study, the 3D pose is estimated directly from the point clouds (as in SPiKE, mAP = 89.19\%). This illustrates the heavy reliance on an intermediate 2D pose in WSM, while SPiKE can accurately regress the 3D pose directly from point cloud sequences alone.

Direct point cloud methods have the advantage of being independent of depth maps, allowing them to handle data from different acquisition methods, such as LiDAR sensors, which produce sparse point clouds. In contrast, 2D-3D lifting methods depend on depth maps (or dense point clouds) to derive an intermediate pose, making them incompatible with sparse point clouds~\cite{zhang2021sequential}.

\subsection{Computational efficiency}

The independence from intermediate 2D keypoint extraction eliminates the need for additional processing of the depth maps, reducing the network complexity and computational needs. Table \ref{tab:runtimes} shows a comparison with 2D-3D lifting methods in terms of inference time (ms) and performance on ITOP (mAP). Since Adapose~\cite{zhang2021sequential} omits its 2D HPE network in their released code, for our comparison, we add the runtime associated with the released code plus the runtime for HRNet, as a representative network for 2D HPE. 

Table \ref{tab:runtimes} shows that SPiKE has a comparable runtime to the 2D-3D lifting modules, but it holds a considerable advantage as it operates independently from a 2D HPE network. This independence provides a significant computational advantage, regardless of the efficiency of the 2D-3D lifting approach employed.

\begin{table}[h]
    \caption{\textcolor{black}{Inference time (ms) per frame and performance (mAP) on ITOP.}}

    \centering
    \begin{tabular}{l>{\centering\arraybackslash}m{3.4cm}>{\centering\arraybackslash}m{2cm}>{\centering\arraybackslash}m{1.5cm}|>{\centering\arraybackslash}m{2cm}}
        \toprule
        Methods & 2D HPE (HRNet) & 2D-3D Lifting & Total & mAP (ITOP) \\
        \midrule
        HRNet+Refinet & 30.34 ms & 5.18 ms & 35.52 ms & 84.2  \\
        AdaPose & 30.34 ms & 13.23 ms & 43.57 ms  & 93.38 \\
        SPiKE (ours) & - & 5.98 ms & 5.98 ms & 89.19 \\
        \bottomrule
    \end{tabular}
    \label{tab:runtimes}
\end{table}

\subsection{Qualitative results}

\begin{figure}[t]
    \centering
    \includegraphics[width=0.95\linewidth]{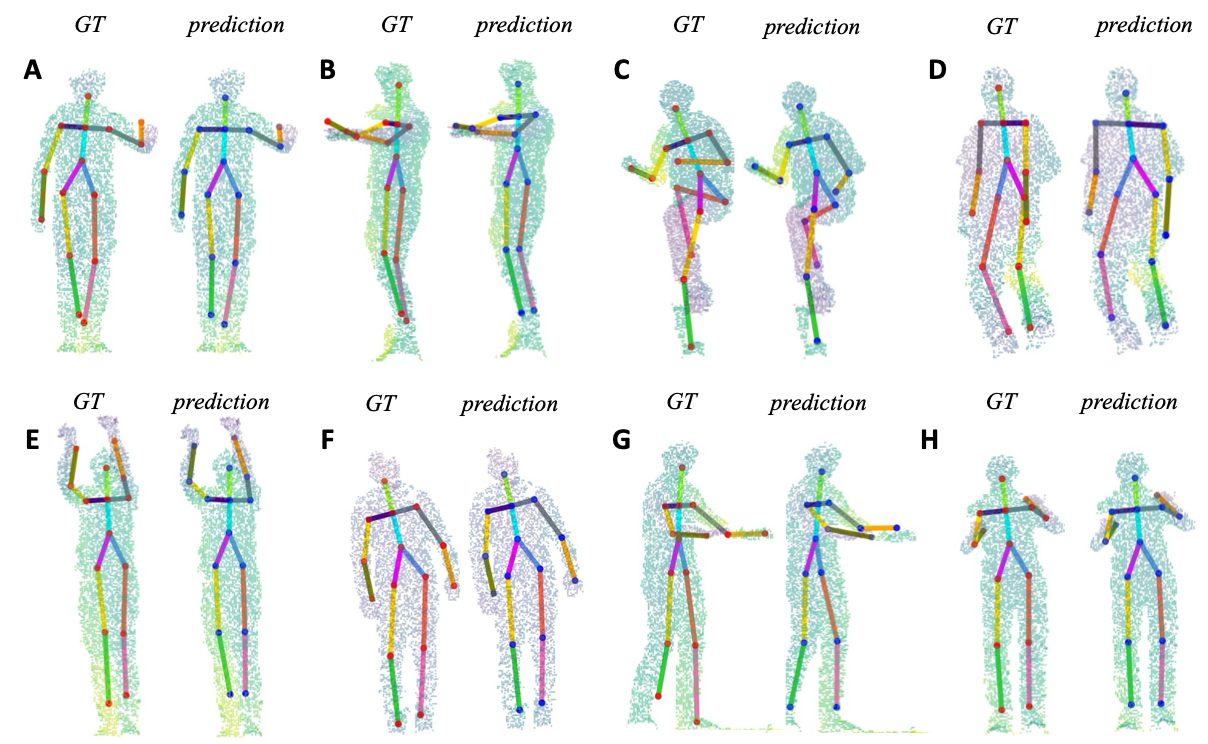} %
    \caption{\textbf{Qualitative results.} Each pair represents the groundtruth skeletons on the left (keypoints in red) and the joints predicted by the model on the right (keypoints in blue). ID top row: A: 0\_01439, B: 2\_00220, C: 1\_00587, D: 3\_02966. ID bottom row: E: 0\_01712, F: 2\_02827, G: 0\_00168, H: 1\_01611.}
    \label{fig:qual}
\end{figure}

Fig. \ref{fig:qual} shows a comparison between ground truth joint coordinates (left, with keypoints in red) and the model's predicted joint positions (right, with keypoints in blue) across a spectrum of poses. This side-by-side view illustrates the model's accuracy in predicting body keypoints. This accurate performance is evident not only in standard poses, such as sample A but also in complex situations where limbs are in motion or partially occluded, as shown in samples B, C, E, G and H. The model is also adept at recognising poses in which the person is turned away from the camera, as shown in sample D, where the colours of the limb joints are inverted from left to right and vice versa, indicating body orientation.

\subsection{Ablations}

We isolate our contributions and algorithmic choices and construct a set of experiments to measure their effect. Specifically, we examine the following aspects of our algorithm: the length of the point cloud sequence, using past or including also future timestamps for pose estimation, and the effect of spatio-temporal convolution instead of spatial convolution to encode local volume features.

\subsubsection{Past vs. past-future timestamps}

When using only past timestamps, the model relies solely on historical data, potentially missing context from future frames, especially in occlusion scenarios. In theory, including future timestamps allows the model to use both past and future information, providing a more complete understanding of the temporal context. 

\begin{figure}[t]
    \centering
    \includegraphics[width=\linewidth]{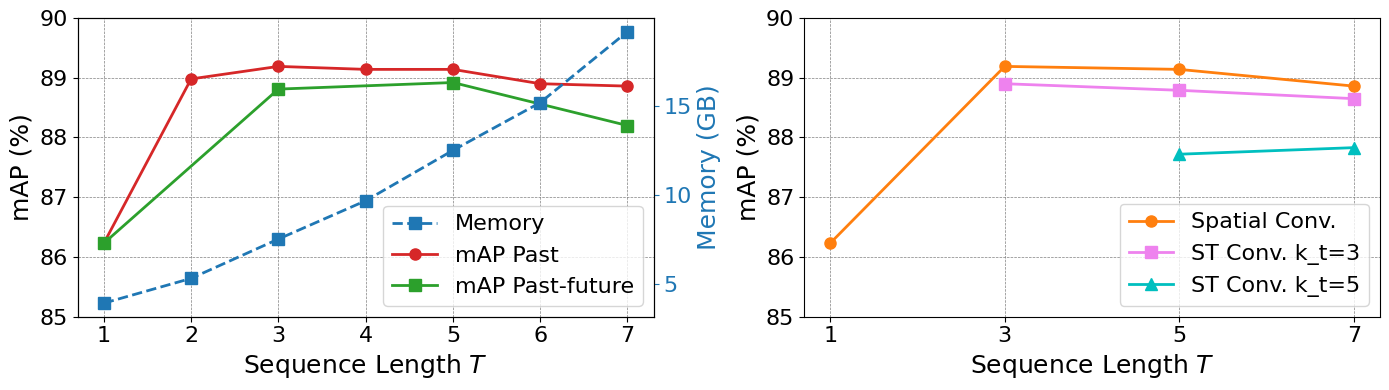} %
    \caption{\textbf{Ablations} Left: Effect on performance (mAP) and running memory (GB) vs. sequence length $T$, using only past or past and future timestamps. Spatial convolutions are employed for this ablation study. Right: Effect on performance (mAP) vs. sequence length $T$ for spatial convolutions, spatio-temporal (ST) convolutions with temporal kernel size $k_t$ = 3 and $k_t$ = 5. For this ablation, only past timestamps are considered.}
    \label{fig:combined_plots}
\end{figure}

However, Fig. \ref{fig:combined_plots} (Left) shows that the performance difference between these approaches is marginal, regardless of sequence length. Thus, while considering future time stamps may theoretically enrich the temporal context, the practical advantage appears to be limited, and models can rely predominantly on past time stamps for efficiency without sacrificing significant performance gains, opening up our method for real-world applications.

\subsubsection{Sequence length}

In action recognition, longer point cloud sequences consistently yield superior results~\cite{fan2021point,fan2022point,fan2021pstnet,liu2019meteornet,wen2022point}. This superiority arises from the uneven distribution of action-related information over time. Consequently, short sequences may overlook critical frames necessary for accurate action inference. 

However, the relationship between sequence length and HPE is not as direct, as the influence of distant timestamps on the current pose may be minimal. Moreover, processing long sequences in HPE requires more memory without necessarily adding significant new information. This phenomenon is illustrated in the left plot of Fig.~\ref{fig:combined_plots} where the mAP peaks at a certain sequence length ($T$~=~3), beyond which the memory requirement continues to increase without a significant improvement in performance. This finding aligns with previous work in HPE from egocentric RGB videos~\cite{wen2023hierarchical}.

\subsubsection{Spatial vs. Spatio-Temporal Convolution}

We compare the effectiveness of spatial against spatio-temporal convolutions~\cite{fan2021point,fan2022point} for feature encoding from local volumes as input to the Transformer. Spatial convolutions maintain the spatial structure within each timestamp, while spatio-temporal convolutions merge information across timestamps, allowing for processing longer sequences. 

Fig. \ref{fig:combined_plots} (Right) shows the performance (mAP) using spatial convolutions and spatio-temporal convolutions with temporal kernel sizes $k_t$ = 3 and $k_t$ = 5. Our findings confirm that spatial convolutions are more effective than spatio-temporal convolutions for HPE due to the fine-grained nature of the task.

\section{Limitations and future work}

Despite the multiple contributions of our work, SPiKE is not without limitations. First, similar to \cite{zhang2020weakly,zhang2021sequential,zhou2020learning}, we apply depth thresholding and clustering to isolate the points belonging to the human. While this strategy is effective for the ITOP dataset, it may not be sufficient for real-world applications. In addition, SPiKE currently focuses on single-human pose estimation and does not address multi-human scenarios. Therefore, future work is needed to address the effective integration of human instance detection as part of the HPE framework. A second line of future work arises from the versatility of SPiKE, which requires only point cloud sequences as input, allowing HPE from point clouds acquired by different sensing devices. Our evaluation is limited to point clouds derived from depth maps, and future work will investigate its performance on datasets of sparse LiDAR point clouds. Finally, future research directions include the adaptation of auto-regressive motion models, such as HuMoR~\cite{rempe2021humor}, for 3D HPE from point cloud sequences.

\section{Conclusion}

We presented SPiKE, a novel approach to 3D HPE from point cloud sequences employing point spatial convolutions and a Transformer architecture to encode spatio-temporal relationships between points along the sequence. We demonstrated that exploiting temporal information by processing sequential point clouds yields superior results compared to treating each timestamp independently. 

To ensure efficient processing while preserving per-timestamp spatial integrity, SPiKE partitions each point cloud of the sequence into local volumes and extracts spatial features through point spatial convolution. Ablation studies confirmed the effectiveness of this strategy and highlighted the superiority of spatial convolutions over spatio-temporal convolutions for HPE.

\textcolor{black}{Experiments on ITOP validated SPiKE's effectiveness, outperforming existing direct approaches with an 89.19\% mAP. Using only point clouds, SPiKE performed comparably to lifting approaches with significantly faster inference.}

Qualitative analysis further underscored SPiKE accuracy across a wide range of poses, including complex scenarios involving occlusion or varying orientations. These findings collectively illustrate the robustness of SPiKE in accurately estimating 3D human poses from point cloud sequences.

\subsubsection{Acknowledgements} This work is supported by the Vienna Science and Technology Fund (AlgoCare - grant agreement No. ICT20-055) and the European Union’s H2020 (VisuAAL - grant agreement No. 861091). The publication reflects the views only of the authors, and the European Union cannot be held responsible for any use which may be made of the information contained therein.
%
%
%
\bibliographystyle{splncs04}
\bibliography{bib_file}

\end{document}